\documentclass[11pt,a4paper]{article}
\usepackage[hyperref]{acl2021}
\usepackage{times}
\usepackage{latexsym}
\usepackage{amsmath,mathtools}
\usepackage{graphicx}
\usepackage{subfigure}
\usepackage{amsfonts}
\usepackage{multirow}
\usepackage{helvet}
\usepackage{booktabs}

\usepackage{microtype}

\aclfinalcopy
\def\aclpaperid{1303} 
\title{Bi-Granularity Contrastive Learning for Post-Training in Few-Shot Scene}

\author{Ruikun Luo, Guanhuan Huang, Xiaojun Quan\thanks{\; Corresponding author.}\\
  School of Computer Science and Engineering, Sun Yat-sen University, China \\
  \texttt{\{luork,~huanggh25\}@mail2.sysu.edu.cn,~quanxj3@mail.sysu.edu.cn} \\}

\date{}

\begin{document}
\maketitle

\begin{abstract}
The major paradigm of applying a pre-trained language model to downstream tasks is to fine-tune it on labeled task data, which often suffers instability and low performance when the labeled examples are scarce.~One way to alleviate this problem is to apply post-training on unlabeled task data before fine-tuning, adapting the pre-trained model to target domains by contrastive learning that considers either token-level or sequence-level similarity. Inspired by the success of sequence masking, we argue that both token-level and sequence-level similarities can be captured with a pair of masked sequences.~Therefore, we propose complementary random masking (CRM) to generate a pair of masked sequences from an input sequence for sequence-level contrastive learning and then develop contrastive masked language modeling (CMLM) for post-training to integrate both token-level and sequence-level contrastive learnings.~Empirical results show that CMLM surpasses several recent post-training methods in few-shot settings without the need for data augmentation. 
\end{abstract}
	
\section{Introduction}
\label{sec:introduction}
	\begin{figure}[t]
		\centering
		\includegraphics[width=0.45\textwidth]{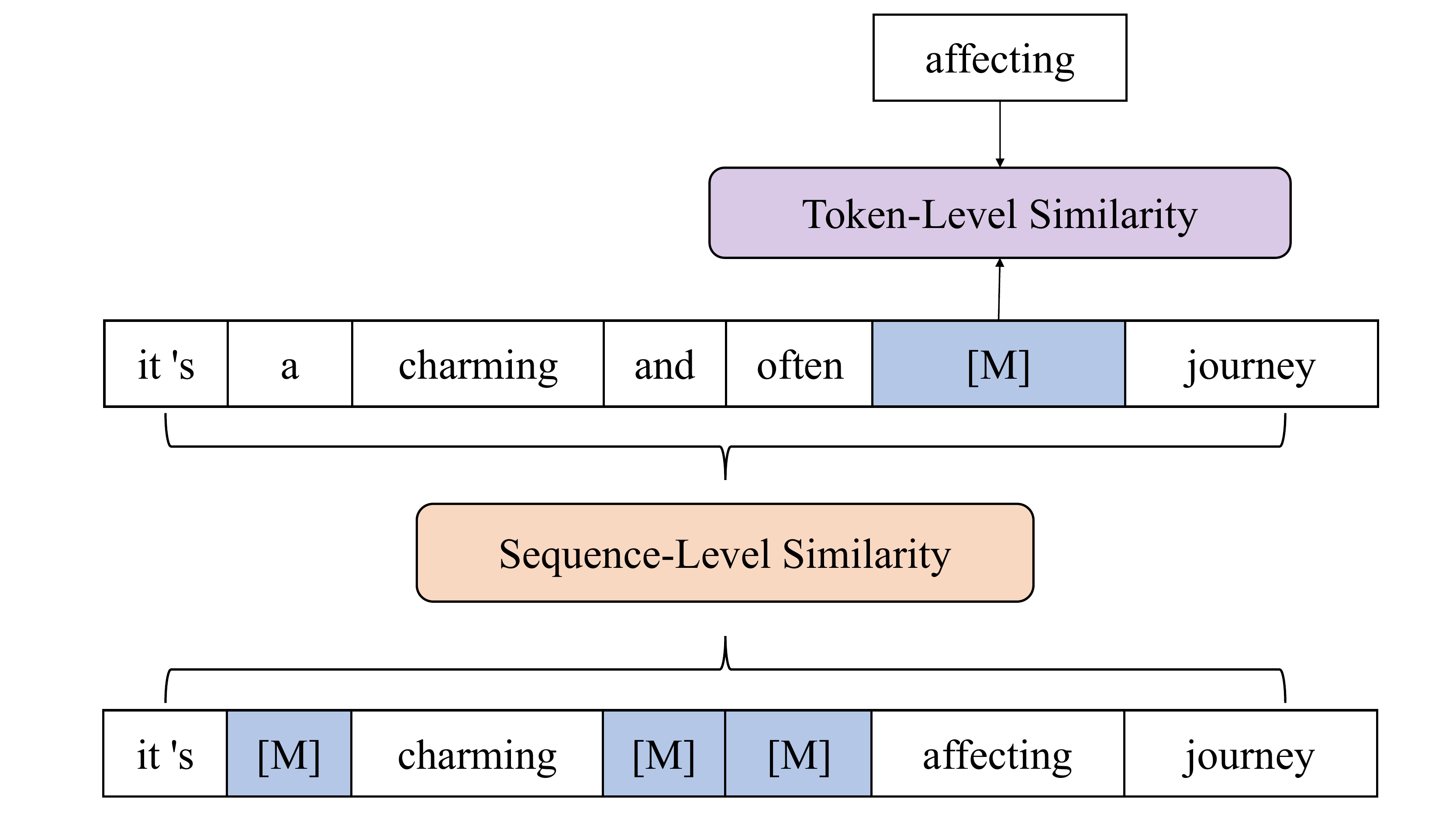}
		\caption{\label{fig:intro}Demonstration of token-level similarity and sequence-level similarity, where the token ``affecting'' is close to the token ``[MASK]'' due to the latter's context. The two sequences with different maskings are also close due to their semantic affinity.}
		\vspace{-0.4cm}
	\end{figure}

	The past few years have seen the rapid proliferation of large-scale pre-trained language models such as GPT \cite{radford2018improving}, BERT \cite{devlin2018bert} and RoBERTa \cite{liu2019roberta}.~These models are generally characterized by pre-training on huge general-domain corpora and then fine-tuning on task-specific labeled examples when applied to a downstream task.~Despite tremendous knowledge obtained from general-domain corpora through pre-training, sufficient labeled examples from the task domain are still needed.~However, collecting them is often infeasible in many scenes, where it tends to be unstable and low-performing when directly fine-tuning these pre-trained models  \cite{zhang2020revisiting,dodge2020fine}. 
	
    Many efforts have been devoted to addressing the above issue. Firstly, it can be relieved by improving the fine-tuning process, such as introducing regularization \cite{jiang-etal-2020-smart,lee2019mixout}, re-initializing top layers \cite{zhang2020revisiting}, and using debiased Adam optimizer \cite{mosbach2020stability}.~Besides, according to empirical results \cite{zhang2020revisiting,mosbach2020stability}, fine-tuning with a small learning rate and more fine-tuning epochs can also improve the situation. Secondly, additional data can be explored, for which two main genres of data might be helpful: labeled examples from related tasks and unlabeled task examples. The former has shown considerable success on the GLUE tasks \cite{phang2018sentence,liu2019multi}, whereas such labeled examples are not always easy to collect. By contrast, the latter is more feasible, especially in scenes where task examples are easy to collect but expensive to label. 

    Contrastive learning \cite{hadsell2006dimensionality} is a recently re-emerged method for leveraging unlabeled data to enhance representation learning. The key to contrastive learning is to capture the similarity between samples. As shown in Figure \ref{fig:intro}, there are two sorts of similarities that can be captured for a natural language sequence: token-level similarity and sequence-level similarity. Masked language model (MLM) \cite{devlin2018bert}, widely adopted in pre-trained language models, can be considered as token-level contrastive learning, as it maximizes the similarity between the ``[MASK]'' token and the original token before masking and minimizes the similarity with other tokens. As for sequence-level similarity, several works \cite{iter-etal-2020-pretraining,giorgi2020declutr,wu2020clear} introduce sequence-level contrastive learning into pre-training. While all these works focus on the pre-training phase, the focal point of this paper is to improve the performance of pre-trained models in downstream tasks through \textit{post-training} especially for scenes where limited labeled data is available.
	
	Speaking of pre-trained language models, \newcite{xu-etal-2019-bert} and \newcite{dontstoppretraining2020} demonstrate improvement in various downstream tasks by training the models with MLM on task examples before fine-tuning, which, following \citet{xu-etal-2019-bert}, is termed \emph{post-training} in this paper even though \newcite{dontstoppretraining2020} term it \emph{adaptive pre-training}. \newcite{fang2020cert} also post-train their model on task examples by contrastive self-supervised learning \cite{pmlr-v119-chen20j}, which pulls together two augmented sentences generated from the same sentence by back-translation \cite{edunov-etal-2018-understanding} while separating those otherwise. However, these works consider either token-level or sequence-level contrastive learning, without integrating them. Moreover, adopting back-translation to generate augmented sentences demands considerable computation and makes the effect of post-training dependent on the translation systems.
	
    To capture both sequence-level and token-level similarities, we propose contrastive masked language modeling (CMLM) to achieve more effective knowledge transfer in post-training and to improve the performance of pre-trained language models in few-shot downstream tasks. For this purpose, we propose complementary random masking (CRM) to generate a pair of masked sequences from a single sequence for both sequence-level and token-level contrastive learnings. We conduct extensive experiments to compare CMLM with several recent post-training approaches, and the empirical results show that CMLM achieves superior or competitive performance in a wide range of downstream tasks. 

    Our contributions can be concluded as follows. First, we propose a new random masking strategy, CRM, to generate a pair of masked sequences favorable to sentence-level contrastive learning. Second, we propose a novel objective, CMLM, for post-training pre-trained language models, which realizes both sequence-level and token-level contrastive learnings on a pair of masked sequences. Third, we compare our approach with several post-training methods and obtain superior or competitive results in few-shot settings.~Lastly, we compare two contrastive learning implementations, SimCLR \cite{pmlr-v119-chen20j} and SimSiam \cite{chen2020exploring}, in pre-trained language models. To our best knowledge, our work is the first effort to implement SimSiam in NLP and compare it with SimCLR.
\vspace{-0.1cm}
\section{Related Works}\vspace{-0.2cm}
\subsection{Pre-trained Language Model}
	Pre-trained language models such as GPT \cite{radford2018improving}, BERT \cite{devlin2018bert} and RoBERTa \cite{liu2019roberta} have become a new paradigm of NLP research, and been successfully applied in a wide range of tasks that used to be thorny.~These models are generally structured with stacks of Transformer \cite{vaswani2017attention} and pre-trained on large-scale unlabeled corpora. Among them, GPT is pre-trained with a unidirectional language modeling objective, and BERT is with masked language modeling (MLM) and next sentence prediction (NSP). In RoBERTa, \newcite{liu2019roberta} turn the static MLM in BERT into a dynamic one and remove the NSP task, and pre-train it more intensively with larger corpora.

    Despite tremendous knowledge learned from unlabeled corpora during pre-training, sufficient labeled task examples are still needed for fine-tuning, which can be challenging for some scenes. Plus, unlabeled task examples are not leveraged when sticking to the pre-training and fine-tuning paradigm.
\vspace{-0.5cm}
\subsection{Contrastive Learning}
\label{sec:cl}
	To take advantage of unlabeled or labeled data more effectively, contrastive learning \cite{hadsell2006dimensionality} is re-emerged recently in computer vision (CV) and natural language processing (NLP). The key to contrastive learning is to pull positive samples together while separating negative samples apart.~The construction of positive and negative sample pairs varies from tasks to tasks. In CV, \newcite{pmlr-v119-chen20j} take the augmented (e.g., by random crop, color distortion, and Gaussian blur) images originated from the same image as positive pairs, and treat those otherwise as negative pairs. \newcite{khosla2020supervised} take the images with the same label as positive pairs and take others as negative. In NLP, MLM in BERT can be viewed as contrastive learning on token level, which takes ``[MASK]'' and its original token before masking as a positive pair and the rest as negative. For sequence level, both \newcite{fang2020cert} and \newcite{wu2020clear} follow \newcite{pmlr-v119-chen20j} to construct the sample pairs. Specifically, \newcite{fang2020cert} utilize back-translation for sequence augmentation while \newcite{wu2020clear} use some easy deformation operations like deletion, reordering and synonym substitution. Besides, \newcite{giorgi2020declutr} construct two spans as a positive pair if they overlap, subsume, or are adjacent. \newcite{gunel2020supervised} introduce supervised contrastive learning proposed by \newcite{khosla2020supervised} into NLP and treat the sequences with the same label as a positive pair. \newcite{li2020cross} augment the cross-entropy loss with a contrastive self-supervised learning term and a mutual information maximization term to deal with the cross-domain sentiment classification task.
\vspace{-0.1cm}
\subsection{Post-training}
\label{sec:post-training}
	Post-training has been broadly applied in downstream tasks. For examples, \newcite{xu-etal-2019-bert} post-train BERT with MLM on task examples to improve the sentiment analysis task.~\newcite{dontstoppretraining2020} further divide post-training into two categories:~domain-adaptive pre-training and task-adaptive pre-training, and evaluate them by extensive experiments.~\newcite{phang2018sentence} fine-tune their model on a related task before fine-tuning on the target task. \newcite{liu2019multi} extend previous work into a multi-task learning fashion. \newcite{fang2020cert} introduce contrastive self-supervised learning (CSSL) \cite{pmlr-v119-chen20j} to perform post-training and name it CSSL Pre-training. 
\vspace{-0.1cm}
\section{Approach}\vspace{-0.2cm}
\label{sec:approach}
	
	\begin{figure*}[t]
		\centering
		\includegraphics[width=0.98\textwidth]{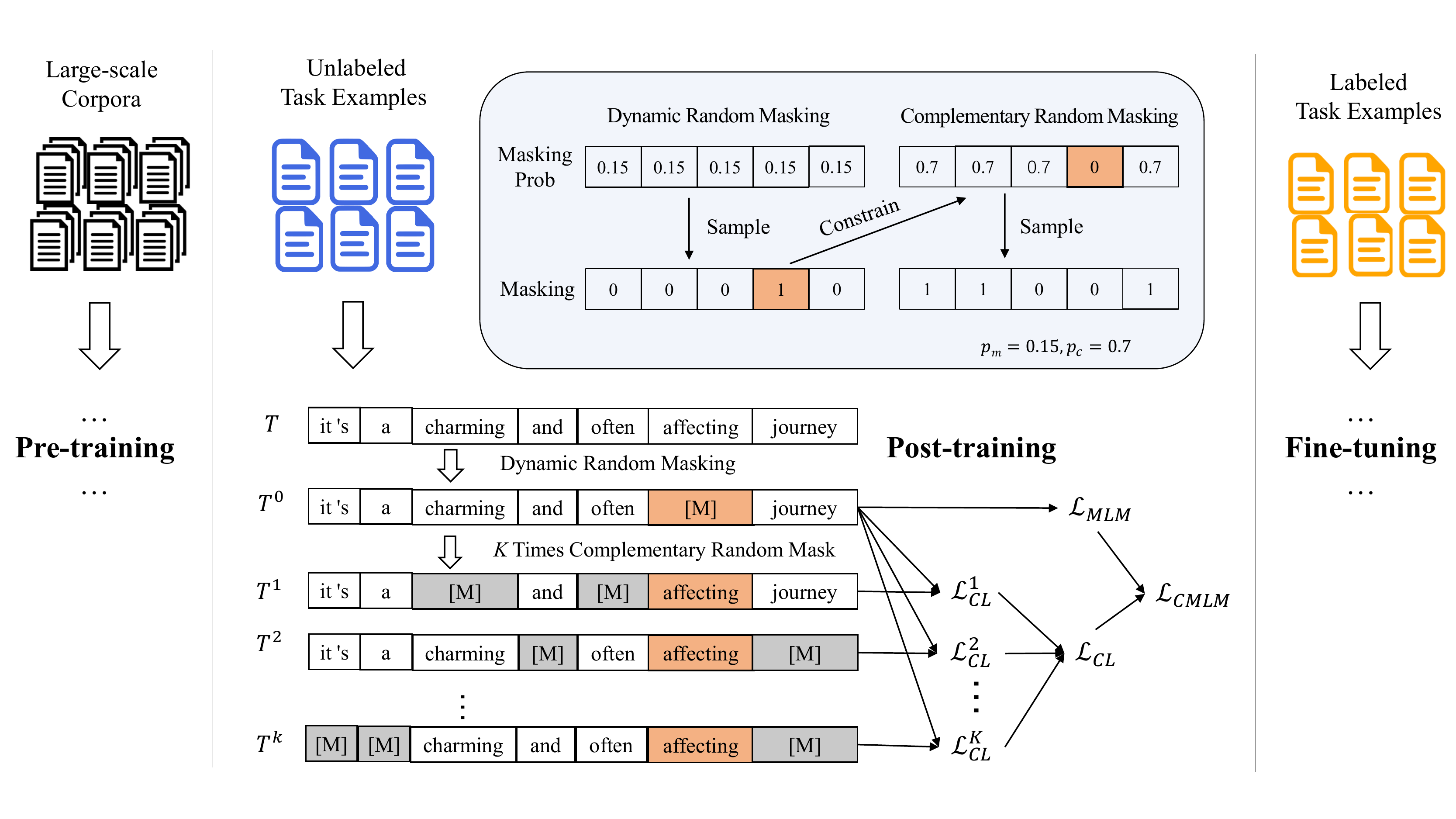}
		\caption{\label{fig:main}Demonstration of complementary random masking (CRM, in the blue box) and contrastive masked language modeling (CMLM).}
	\end{figure*}
	
\subsection{Dynamic Random Masking}
\label{sec:DRM}
	 Masked language modeling (MLM) is firstly applied in BERT \cite{devlin2018bert}, where some of the tokens in the input sequence are selected and replaced by a special token ``[MASK]''. BERT uniformly selects 15\% of the input tokens for replacement, and among the selected tokens, 80\% are replaced with ``[MASK]'', 10\% are left unchanged, and 10\% are replaced by a randomly selected vocabulary token. In the original implementation of BERT, random masking and replacement are performed once in the beginning, and the sequences are kept unchanged through pre-training. \newcite{liu2019roberta} transform this static masking strategy into dynamic random masking (DRM) by generating a masking pattern every time a sequence is fed. That is to say, given an input sequence $T = \{t_1, t_2, ..., t_N\}$, the probability of each token being selected is determined by $p_m$, which is fixed to 15\% in BERT and RoBERTa.
    \begin{equation}
    \label{eq:mp}
    	P_{DRM}(t_n) = p_m,\ n \in [1,N]
    \end{equation}
    
\subsection{Complementary Random Masking}
\label{sec:CRM}
	It is straightforward to come up with an idea that generates a pair of masked sequences from a single sequence by random masking and applies MLM on each masked sequence to capture token-level similarity, and then perform sequence-level contrastive learning between the two sequences to capture sequence-level similarity. However, it faces a dilemma when applying this idea: setting a small $p_m$~would make the pair of masked sequences too similar and make the contrastive learning loss drop to 0 quickly, harming sequence-level contrastive learning.~On the other hand, setting a large $p_m$ would make each masked sequence collapsed, making it hard for the model to recover the original tokens from ``[MASK]'' based on the context, which in turn harms token-level contrastive learning.
	
    To address this issue, we decouple the pair of masked sequences, denoted by $T^0$ and $T^1$, and assign them different masking probabilities $p_m$ and $p_c$. Specifically, we obtain $T^0$ with a small masking probability $p_m$ and $T^1$ with a larger probability $p_c$. Moreover, to avoid a single word being masked by both sequences, which cripples their relevance, we propose complementary random masking (CRM) to generate a pair of complementary masked sequences which maintain a complementary relationship. Concretely, in CRM, we first generate $T^0 = \{t_1^0, t_2^0, ..., t_N^0\}$ with DRM  \cite{liu2019roberta} from the original sequence $T$, and generate $T^1$ with an extra constraint: the masking probability $P_{CRM}(t_n)$ will be set to $p_c$ if and only if $t_n^0$ has not been selected in $T^0$. Otherwise, it will be set to 0. The process of CRM is described in Figure \ref{fig:main}. 
    \begin{gather}
		\label{eq:cp}
		\mathclap{
		P_{CRM}(t_n) = \begin{cases}
			p_c, & \text{$t_n^0$ was not selected in~}T^0\\
			0, & \text{otherwise} \qquad
		\end{cases}}
	\end{gather}

    CRM is aimed to generate a complementary pair of masked sequences: If $p_c=1$, all tokens that are not selected in $T^0$ will be selected in $T^1$. Reducing $p_c$ can soften this complementary relationship and make the two sequences overlap increasingly.
	
\subsection{Contrastive Masked Language Modeling}
\label{sec:CMLM}
	We propose contrastive masked language modeling, CMLM, based on CRM to realize domain transfer by masked language modeling (MLM) and sequence-level contrastive learning (CL) with pairs of masked sequences. The framework of CMLM is shown in Figure \ref{fig:main}, which is described as follows.
	
	Given a batch $\mathcal{T} = \{T_1, T_2, ..., T_B\}$ of input sequences, we firstly apply dynamic random masking (DRM) on each sequence $T_b$ to generate a masked sequence $T_b^0$, and then apply CRM $K$ times to generate $T_b^1, T_b^2, ..., T_b^K$ based on $T_b^0$ and $T_b$:
	\begin{gather}
		\label{eq:mask}
		T_b^{0} = {\rm{DRM}}(T_b),\ b \in [1,B]\\
		T_b^{k} = {\rm{CRM}}(T_b, T_b^{0}),\ k \in [1,K]
	\end{gather}
    After obtaining $K+1$ masked sequences from each sequence $T_b$ in $\mathcal{T}$, we then compute their representations $H_b^k \in \mathbb{R}^{N \times d}$ by using an encoder, where $d$ is the hidden size of the encoder:
	\begin{gather}
		\label{eq:encoder}
		H_b^k = {\rm{Encoder}}(T_b^k),\ k \in [0,K]
	\end{gather}
    Even though our approach is model-agnostic, in this paper we focus on the Transformer-based pre-trained language model RoBERTa, which is an enhanced version of BERT. Therefore, we employ RoBERTa to implement the $\rm{Encoder}(\cdot)$ function.
	
	To capture token-level similarity, We apply MLM on $H_b^0$ as \newcite{devlin2018bert} and \newcite{liu2019roberta}, and compute the loss as follows:
	\begin{gather}
		\mathcal{L}_{MLM} = \frac{1}{B}\sum^{B}_{b=1}{\rm{MLM}}(H_b^0)
	\end{gather}
	
	To capture sequence-level similarity, we apply contrastive learning on each $H^k_b$ and $H^0_b$, and obtain the loss term $\mathcal{L}_{CL}$. We compare two different implementations of contrastive learning: SimCLR \cite{pmlr-v119-chen20j} and SimSiam \cite{chen2020exploring}. For SimCLR, $\mathcal{L}_{CL}$ can be calculated as:
	\begin{gather}
		\label{eq:simclr}
		\mathclap{
		\mathcal{L}_{CL}\hspace{-0.1cm}=\hspace{-0.1cm}-\frac{1}{K\cdot B}\hspace{-0.1cm}\sum_{k=1}^{K}\hspace{-0.1cm}\sum_{b=1}^{B}\hspace{-0.1cm}{\rm{log}}\frac{e^{{\rm{sim}}(H^k_b, H^0_b)/\tau}}{\sum_{i=1}^B e^{{\rm{sim}}(H_i^k, H_b^0) / \tau}}} \quad
	\end{gather}
	where $\tau$ is a temperature parameter.
	
	Following \newcite{gunel2020supervised}, we take the first token representation $h_b^k \in R^d$ of $H_b^k$ to calculate the similarity between $H^k_i$ and $H^0_j$ as follows.
	\begin{gather}
		\label{eq:sim}
		{\rm{sim}}(H^k_i, H^0_j) = \frac{h^k_i}{||h^k_i||_2} \cdot \frac{h^0_j}{||h^0_j||_2}
	\end{gather}

    SimSiam is similar to SimCLR except without using negative pairs and has a negative loss value. To be consistent with $\mathcal{L}_{MLM}$ and $\mathcal{L}_{CL}$, we define the loss function of SimSiam as follows:
	\begin{gather}
		\label{eq:simsiam}
		\mathclap{
		\mathcal{L}_{CL}\hspace{-0.1cm} = \hspace{-0.1cm}\frac{1}{K\cdot B}\sum_{k=1}^{K}\hspace{-0.05cm}\sum_{b=1}^{B}e^{-\frac{1}{2}(D(z_b^k, h_b^0)+D(z_b^0, h_b^k))}}\quad
	\end{gather}
    where $z^k_b$ and $D(z,h)$ are defined as:
	\begin{gather}
		z_b^k = W_2 \cdot {\rm{gelu}}(W_1 \cdot h_b^k)\\
		D(z,h) = {\rm{sim}}(z, stopgrad(h)).
	\end{gather}	
    Here, $W_1,W_2 \in R^{d \times d}$ are learnable parameters, ${\rm{sim}}(\cdot)$ is similar to Equation \ref{eq:sim}, and ${\rm{Stopgrad}}(\cdot)$ is a stop-gradient operation which is crucial for SimSiam \cite{chen2020exploring}.
	
	Finally, we combine $\mathcal{L}_{MLM}$ and $\mathcal{L}_{CL}$ for contrastive masking language modeling:
	\begin{gather}
		\label{eq:alpha}
		\mathcal{L}_{CMLM} = \mathcal{L}_{MLM} + \alpha \cdot \mathcal{L}_{CL}
	\end{gather}
	where $\alpha$ is a tunable hyper-parameter.
	
\subsection{Relationship to Existing Approaches}
	Among existing approaches, the closest one to ours is CSSL Pre-training  \cite{fang2020cert}. We can implement CSSL Pre-training by slightly modifying Equation \ref{eq:mask} and \ref{eq:alpha} to following ones:
	\begin{gather}
		T_b^k = {\rm{back}}(T_b),\ k \in [0,1]\\
		\mathcal{L}_{CSSL} = \mathcal{L}_{CL}
	\end{gather}
    where ${\rm{back}}(T)$ means back-translation of $T$. And other equations stay the same. By comparing these equations, we note that CMLM can be considered as: (1) replacing back-translation with CRM, which not only reduces the computational cost but also prevents the model from depending on the translation systems; (2) adding $\mathcal{L}_{MLM}$ to implement token-level contrastive learning, which is shown to be crucial in Section \ref{sec:mlm_critical}; (3) easily extending one pair of positive samples to $K$ pairs, which can be attributed to the nature of random masking.
	
	As for the loss term, both \newcite{giorgi2020declutr} and \newcite{wu2020clear} use similar terms to ours for pre-training: $\mathcal{L}_{total} = \mathcal{L}_{MLM} + \mathcal{L}_{CL}$, where $\mathcal{L}_{MLM}$ captures token-level similarity and $\mathcal{L}_{CL}$ captures sequence-level similarity. The main difference between these methods and ours is that we use differently masked sequences from the same sequence as a positive pair, while \newcite{giorgi2020declutr} use position-related segments (overlapping, adjacent or subsumed) and \newcite{wu2020clear} use sequences by different deformations as the positive pair.
	
\section{Experiment}\vspace{-0.1cm}
\subsection{Tasks: GLUE}\vspace{-0.1cm}
\label{sec:dataset}
	We evaluate our model on the GLUE benchmark \cite{wang-etal-2018-glue}, which contains 9 natural language understanding tasks that can be divided into three categories:~(1) single sentence tasks: CoLA and SST-2; (2) similarity and paraphrase tasks:~MRPC, QQP, and STS-B; (3) inference tasks: MNLI, QNLI, RTE, and WNLI. All of them are classification tasks except STS-B, so we eliminate it to focus on the classification tasks. WNLI has a small development set (70 examples) and is also ignored.~MNLI contains two evaluation sets. One, denoted as MNLI, is from the same sources as the training set, and the other, denoted as MNLI-MM, is from different sources than the training set.
	
    To simulate few-shot scenes of different degrees, we randomly select 20, 100, and 1000 examples respectively from these tasks as our training sets following recent work \cite{gunel2020supervised}. For each subset in each task, we sample 5 times with replacement and obtain 15 training sets for each task. As for the development set and test set, we randomly select 500 examples from the original development set as our development set and take the remaining as our test set. Since QQP contains too many examples (40k) in the original development set, we randomly select 2000 from the remaining examples after sampling our development set as our test set. Note that all the 15 training sets in each task share the same development and test sets.
\vspace{-0.15cm}	
\subsection{Model: RoBERTa}\vspace{-0.05cm}
	As mentioned above, we take RoBERTa to implement our encoder in Equation \ref{eq:encoder}.~The base version of RoBERTa, Roberta-base, which contains 12 Transformer blocks with 12 self-attention heads, is employed. All the blocks have the same hidden size 768. The input sequence is either a segment or two segments separated by a special token ``[$\backslash$s]'', while ``[s]'' is always the first token. We take the implementation and pre-trained weights from Huggingface Transformers library \cite{wolf-etal-2020-transformers}.
	
	\begin{table*}
	\centering
	\scalebox{0.7}{
		\begin{tabular}{lcccccccc}
            \toprule
            \multicolumn{1}{l}{} & \textbf{CoLA}   & \textbf{SST-2}  & \textbf{MNLI}   & \textbf{MNLI-MM} & \textbf{QNLI}   & \textbf{RTE}    & \textbf{MRPC}   & \textbf{QQP}    \\ \hline
            \textbf{Metric}      & mcc             & acc             & acc             & acc              & acc             & acc             & acc             & acc             \\ \hline
            \multicolumn{9}{c}{data size = 20}                                                                                                                                    \\ \hline
            FT                   & 0.0751$\pm$6.26          & 0.6604$\pm$6.38          & 0.3578$\pm$1.70          & 0.3652$\pm$1.96           & 0.6163$\pm$3.46          & 0.5281$\pm$4.07          & \textbf{0.6747$\pm$2.23} & 0.6777$\pm$3.67 \\
            SCL                  & \textbf{0.1105$\pm$8.30} & 0.6964$\pm$6.24          & \textbf{0.3684$\pm$2.85} & \textbf{0.3751$\pm$3.41}  & 0.6191$\pm$3.55          & 0.5082$\pm$7.75          & 0.6631$\pm$1.15          & \textbf{0.6947$\pm$2.35}\\
            CSSL                 & 0.0795$\pm$4.13          & 0.6609$\pm$5.98          & 0.3640$\pm$2.06          & 0.3686$\pm$2.69           & 0.6064$\pm$3.26          & 0.5264$\pm$7.10          & 0.6638$\pm$1.64          & 0.6514$\pm$2.99\\
            TAPT                 & 0.0860$\pm$7.64          & 0.7326$\pm$4.99          & 0.3616$\pm$2.12          & 0.3689$\pm$2.40           & 0.6146$\pm$3.60          & 0.5437$\pm$5.06          & 0.6552$\pm$1.64          & 0.6584$\pm$3.33\\
            {CMLM (ours)} & 0.0902$\pm$8.65          & \textbf{0.7371$\pm$5.64} & 0.3633$\pm$2.15          & 0.3701$\pm$2.57           & \textbf{0.6231$\pm$3.60} & \textbf{0.5437$\pm$4.57} & 0.6586$\pm$1.41          & 0.6541$\pm$3.89\\ \hline
            \multicolumn{9}{c}{data size = 100}                                                                                                                                   \\ \hline
            FT                   & 0.2176$\pm$7.78          & 0.8405$\pm$2.71          & 0.4361$\pm$2.50          & 0.4526$\pm$2.94           & 0.6820$\pm$2.56          & 0.5879$\pm$4.82          & \textbf{0.7099$\pm$1.72} & \textbf{0.7511$\pm$1.72}\\
            SCL                  & 0.2467$\pm$5.46          & 0.8455$\pm$1.38          & 0.4499$\pm$3.30          & 0.4627$\pm$3.84           & 0.6765$\pm$2.47          & 0.5835$\pm$6.41          & 0.7063$\pm$1.45          & 0.7461$\pm$1.63\\
            CSSL                 & 0.1719$\pm$7.90          & 0.8401$\pm$1.71          & 0.4185$\pm$2.98          & 0.4298$\pm$3.55           & 0.6701$\pm$1.89          & 0.5532$\pm$5.10          & 0.7038$\pm$1.58          & 0.7274$\pm$1.85\\
            TAPT                 & 0.2626$\pm$6.03          & 0.8496$\pm$2.52          & 0.4508$\pm$2.60          & 0.4682$\pm$2.80           & 0.6970$\pm$1.63          & 0.6095$\pm$6.60          & 0.6987$\pm$1.77          & 0.7429$\pm$2.13\\
            {CMLM (ours)} & \textbf{0.2663$\pm$6.97} & \textbf{0.8525$\pm$1.95} & \textbf{0.4530$\pm$2.75} & \textbf{0.4683$\pm$3.00}  & \textbf{0.6980$\pm$1.67} & \textbf{0.6147$\pm$6.36} & 0.6933$\pm$1.90          & 0.7479$\pm$2.16\\ \hline
            \multicolumn{9}{c}{data size = 1000}                                                                                                                                  \\ \hline
            FT                   & 0.4216$\pm$3.13          & 0.8996$\pm$0.97          & 0.7048$\pm$1.19          & 0.7168$\pm$1.17           & 0.7681$\pm$1.07          & 0.7472$\pm$2.50          & 0.8223$\pm$1.22          & \textbf{0.7934$\pm$0.90}\\
            SCL                  & 0.2758$\pm$11.94          & 0.8991$\pm$1.04          & 0.5020$\pm$3.81          & 0.5089$\pm$4.09           & 0.7449$\pm$1.22          & 0.7100$\pm$5.44          & 0.7157$\pm$6.83          & 0.7853$\pm$0.77 \\
            CSSL                 & 0.4069$\pm$3.53          & 0.8993$\pm$1.38          & 0.6900$\pm$1.37          & 0.7048$\pm$1.39           & 0.7760$\pm$0.97          & 0.7082$\pm$5.30          & 0.8261$\pm$1.70          & 0.7881$\pm$1.00 \\
            TAPT                 & 0.4362$\pm$3.46          & 0.9016$\pm$0.70          & 0.7074$\pm$1.79          & 0.7203$\pm$1.68           & 0.7689$\pm$0.72          & 0.7524$\pm$4.06          & 0.8214$\pm$1.14          & 0.7890$\pm$0.82 \\
            {CMLM (ours)} & \textbf{0.4374$\pm$2.06} & \textbf{0.9023$\pm$0.88} & \textbf{0.7110$\pm$2.00} & \textbf{0.7247$\pm$1.84}  & \textbf{0.7719$\pm$0.91} & \textbf{0.7610$\pm$3.28} & \textbf{0.8223$\pm$0.82} & 0.7891$\pm$0.90 \\ \bottomrule
        \end{tabular}
    }
	\caption{\label{tab:few}Results on the GLUE benchmark with 20, 100 and 1000 training examples, respectively, and compared with baseline (FT: \cite{liu2019roberta}) and several recent post-training or contrastive learning methods (SCL \cite{gunel2020supervised}, CSSL \cite{fang2020cert}, TAPT \cite{dontstoppretraining2020}). Unit of standard deviation is $10^{-2}$. }
	\vspace{-0.1cm}
    \end{table*}
	
	\vspace{-0.15cm}
\subsection{Training Details}	\vspace{-0.05cm}
\label{sec:train_detail}
For the fine-tuning of all approaches to be reported below, unless otherwise specified, we use AdamW \cite{loshchilov2018decoupled} with a learning rate of 1e-5 and epochs of 350, 100, 10 for subsets sized 20, 100, 1000, respectively. This setting is based on previous empirical results \cite{zhang2020revisiting,mosbach2020stability}, which show that fine-tuning with a small learning rate and more epochs stabilizes the performance of a model in few-shot scenes. We set the batch size to 16 and dropout rate to 0.1, and save model parameters every 100 update steps and pick the best based on validation.
	
    For post-training of CMLM, we apply AdamW with a learning rate of 1e-5 and epochs of 200, 50, 5 for subsets sized 20, 100, 1000, respectively. For a fair comparison with other approaches, we set $K$ in Equation \ref{eq:mask} to 1 and the batch size to 8, where the maximum GPU memory usage is approximately equal to that of fine-tuning. For the implementation of $\mathcal{L}_{CL}$, we choose SimSiam for it consumes less computation.~For $p_m$ in Equation \ref{eq:mp}, we follow \cite{liu2019roberta} and set it to 0.15. We conduct a grid-based search for hyper-parameters with $\alpha \in \{0.01, 0.1, 0.3, 0.5, 0.7, 1\}$ (Equation \ref{eq:alpha}) and $p_c \in \{0.1, 0.3, 0.5, 0.7, 0.9\}$ (Equation \ref{eq:cp}), and find that the combination of $\alpha = 0.5$ and $p_c = 0.7$ performs the best on the development set.
	
    For the baselines to be introduced below, we follow the same fine-tuning and post-training settings as our CMLM, with only several method-specific hyper-parameters unchanged. 
	
	\vspace{-0.15cm}
\subsection{Baseline Approaches}\vspace{-0.05cm}
    As mentioned in Section \ref{sec:cl} and \ref{sec:post-training}, there have been works trying to add extra loss terms in fine-tuning or to insert a post-training phase in between pre-training and fine-tuning.~To make a comprehensive comparison, we employ the following approaches as our baselines: (1) fine-tuning (FT) \cite{liu2019roberta}, which directly fine-tunes a model with cross-entropy loss; (2) fine-tuning with SCL (SCL) \cite{gunel2020supervised}, which fine-tunes a model with cross-entropy loss and supervised contrastive loss; (3) post-training with CSSL (CSSL) \cite{fang2020cert}, which post-trains a model with contrastive self-supervised learning loss; (4) post-training with MLM (TAPT) \cite{dontstoppretraining2020}, which post-trains a model with MLM loss and is equal to CMLM when $\alpha=0$. Comparing with recent works \cite{fang2020cert,gunel2020supervised} that take only the conventional either BERT or RoBERTa as their baseline, we consider a few more baselines to obtain more conclusive results.
	\vspace{-0.15cm}
\subsection{Evaluation Details}\vspace{-0.05cm}

    In few-shot scenes, the distribution of the training set may deviate from the test set seriously. \newcite{gunel2020supervised} pick the top-3 results from all combinations of training sets and model seeds for each task. Differently, for each data size of 20, 100, and 1000 described in Section \ref{sec:dataset}, we train our model with random seeds \{31, 42, 53\} for the 5 training subsets, and calculate the mean and standard deviation of the 15 test results. We assume this is a better way to evaluate the overall effect of our model. 
    
    \vspace{-0.15cm}
\subsection{Few-Shot Results}\vspace{-0.05cm}
    In Table \ref{tab:few}, we report our few-shot results on the GLUE tasks with 20, 100, and 1000 training examples, respectively.~Five observations can be made from the table.~First, CMLM obtains superior performance on the datasets with 100 and 1000 examples, surpassing the baselines in 13 of 16 tasks. Since we use the same hyper-parameters for these approaches and report the average results over 3 random seeds and 5 randomly sampled training sets, these results are convincing.~Second, on the dataset with 20 training examples, CMLM only surpasses the other approaches in 3 of 8 tasks. Training a model with only 20 examples is very unstable, and the test results of the baseline approaches indeed show large deviations across different training sets. Third, we find that post-training with only $\mathcal{L}_{MLM}$ \cite{xu-etal-2019-bert,dontstoppretraining2020} can achieve competitive results with the baselines, showing the effectiveness of this widely-used approach. Fourth, SCL \cite{gunel2020supervised} has extremely poor performance on CoLA, MNLI, and MRPC when the data size is 1000, which is beyond our expectation. In the original paper, the authors of SCL only report the top-3 results from combinations of model seeds and train sets. So we speculate this under-performance might come from the instability of SCL in few-shot settings. Fifth, CSSL \cite{fang2020cert} performs even worse than FT when the data size is either 20 or 100 but achieves competitive results when the data size is 1000. CSSL is designed for full-size GLUE tasks and might not be suitable for the few-shot scenes.
    
\subsection{Full-Size Results}
	To verify whether post-training with CMLM can still achieve desirable results when sufficient labeled examples are available, we conduct experiments on the RTE (2.5k), MRPC (3.7k), CoLA (8.5k), SST-2 (67k), and QNLI (106k) tasks with their full-size training sets. We set the learning rate to 3e-5 for both post-training and fine-tuning and set the epoch to 3. Other hyper-parameters remain the same as in Section \ref{sec:train_detail}. Experiment results are shown in Table \ref{tab:full}, from which we can note that CMLM maintains its superiority on RTE and CoLA but fails on MRPC, SST-2, and QNLI. TAPT performs better on tasks with more training examples, which can be explained by the better generalizability of token-level representation, though it demands more training steps to learn well. Note that CMLM is specifically proposed for few-shot settings, so the experiments in the full-size setting are only to evaluate it from different perspectives and make a comprehensive comparison with baselines. 
	\begin{table}[h]
		\centering
		\scalebox{0.75}{
			\begin{tabular}{lccccc}
				\toprule
				\multicolumn{1}{l}{} & \textbf{RTE}    & \textbf{MRPC}   & \textbf{CoLA}   & \textbf{SST-2}  & \textbf{QNLI} \\ \hline
				metric               & acc             & acc             & mcc             & acc             & acc           \\ \hline
				data-size            & 2.5k            & 3.7k            & 8.5k            & 67k             & 106k          \\ \hline
				FT                   & 0.7403          & 0.8623          & 0.5552          & 0.9319          & 0.9043        \\
				SCL                  & 0.6753          & 0.7393          & 0.5329          & \textbf{0.9373} & 0.9002        \\
				CSSL                 & 0.6623          & \textbf{0.8713} & 0.5217          & 0.9310          & 0.8904        \\
				TAPT                 & 0.7403          & 0.8541          & 0.5519          & 0.9355          & \textbf{0.9063} \\
				{CMLM (ours)} & \textbf{0.7446} & 0.8574          & \textbf{0.5714} & 0.9310          & 0.9039        \\ \bottomrule
			\end{tabular}
		}
		\caption{\label{tab:full}Results on the RTE, MRPC, CoLA, SST-2 and QNLI tasks with full-size training sets, and average results over 3 random seeds are reported.}
			\vspace{-0.4cm}
	\end{table}
	
\subsection{Additional Unlabeled Examples}
\begin{figure}[t]
		\centering
			\includegraphics[width=0.48\textwidth]{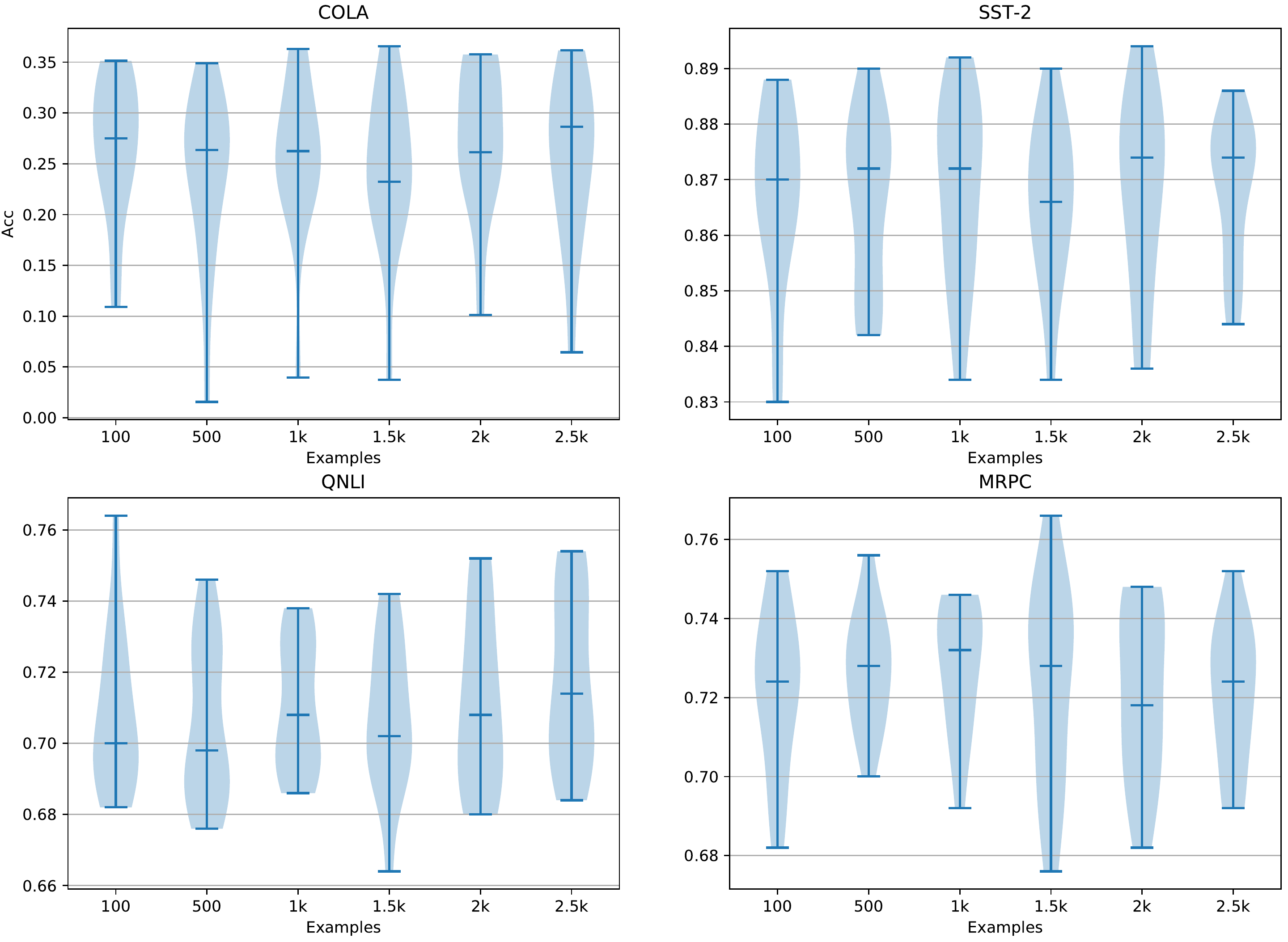}
		\caption{\label{fig:curve}Results of our model on development sets with increasing numbers of unlabeled examples.}
		\label{fig-curve-examples}
			\vspace{-0.1cm}
	\end{figure}

    We consider the scene where additional unlabeled task examples are provided. We evaluate CMLM on CoLA, SST-2, QNLI, and MRPC with 100 labeled examples for fine-tuning and increase unlabeled examples from 100 to 2500 for post-training. We depict the results on the development sets in Figure \ref{fig-curve-examples}, from which two observations can be made. First, the performance generally increases with the number of unlabeled examples grows, showing the helpfulness of unlabeled task examples, which is also confirmed by \newcite{dontstoppretraining2020}. Second, there are certain fluctuations in the results. We assume they come from the random nature of these additional unlabeled examples, which are sampled from a much larger training set and might severely deviate from the original training set. Moreover, we should acknowledge that adding more unlabeled examples in post-training gains limited improvement compared with adding more labeled examples in fine-tuning. Labeled examples are more treasurable in classification learning.


\subsection{Hyper-parameter $K$}
\begin{figure}[t]
\centering
	\includegraphics[width=0.48\textwidth]{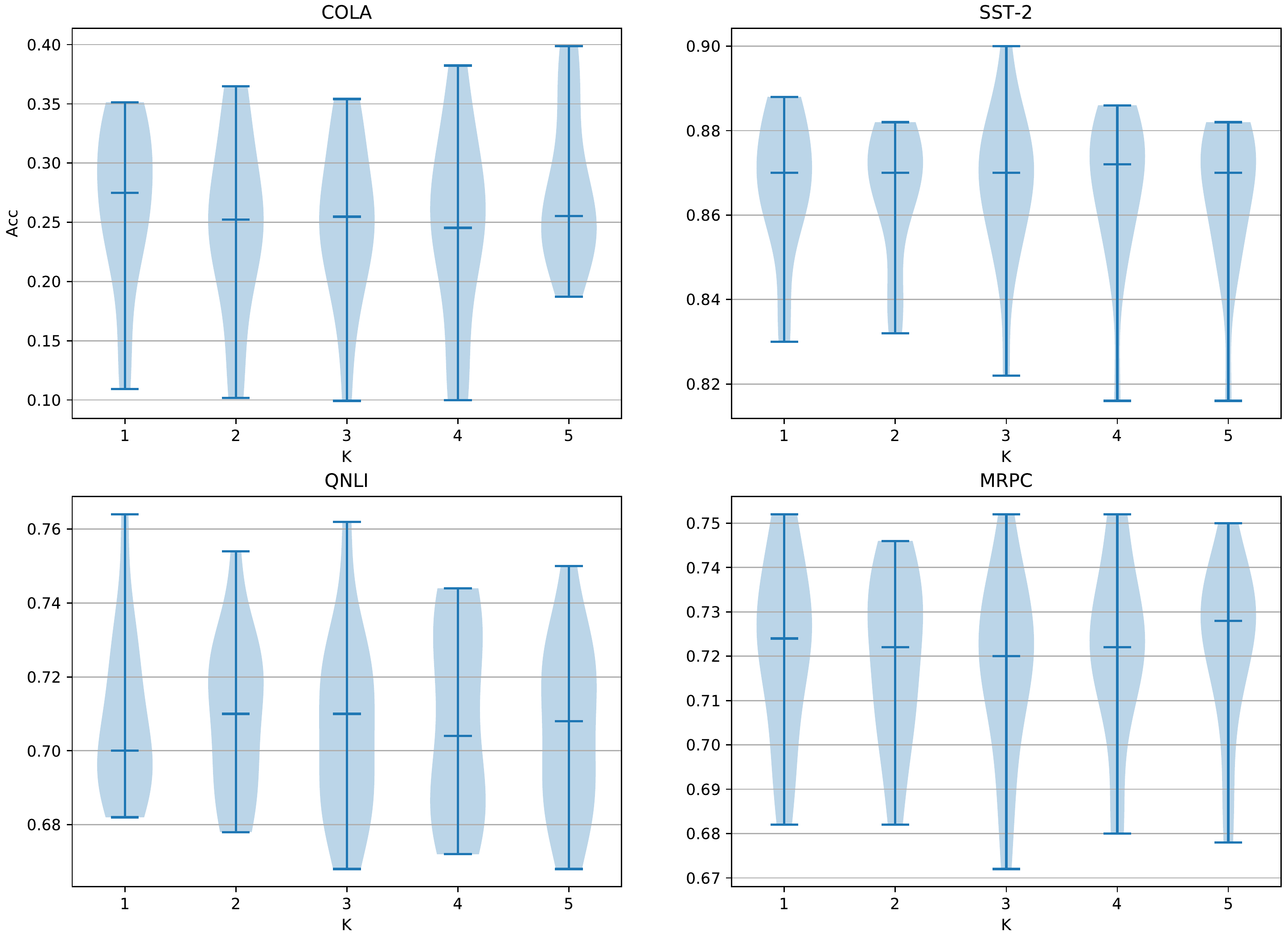}
	\caption{Results of our model on development sets while the parameter $K$ changes.}
		\label{fig-curve-k}
					\vspace{-0.1cm}
	\end{figure}

    We evaluate whether increasing $K$ (in Equation \ref{eq:simclr}) can lead to improvement of our model in the few-shot setting. We evaluate CMLM on the CoLA, SST-2, QNLI, and MRPC tasks with 100 training examples by increasing $K$, and the results are depicted in Figure \ref{fig-curve-k}. Similar to increasing unlabeled examples, the performance slightly improves on the 5 tasks but has some fluctuations, which is within our expectation. Intuitively, exposing the model to different forms of masked sequences can better reflect the distribution of examples sampled from a large training set, but cannot narrow the deviation between these examples and the original train set.

\section{Ablation Studies}
\subsection{SimCLR vs SimSiam}
	
    As described above, $\mathcal{L}_{CL}$ can be implemented by either Equation \ref{eq:simclr} or Equation \ref{eq:simsiam}, although we implement the latter to conduct the above experiments for its less computational cost. According to  \newcite{chen2020exploring}, SimSiam performs better than SimCLR on ImageNet \cite{5206848}. It is thus interesting to verify whether the same holds in our situation. We compare SimSiam (CMLM) and SimCLR (w/ SimCLR) on SST-2, CoLA, QNLI and RTE, and the results are reported in Table \ref{tab:ablation}. From the results, we cannot easily conclude which one is better due to their comparable performances, yet further investigation is beyond the scope of this paper. However, we prefer SimSiam due to it consumes less computation and is easier to implement.

\subsection{Are MLM \& CL Critical for CMLM?}
\label{sec:mlm_critical}
    One of the improvements of CMLM over previous works is combining $\mathcal{L}_{MLM}$ and $\mathcal{L}_{CL}$ to implement both token-level and sequence-level contrastive learnings. Here, we verify how the bi-granularity contrastive learnings contribute to the performance differently.~We remove $\mathcal{L}_{MLM}$ and $\mathcal{L}_{CL}$ alternatively from $\mathcal{L}_{CMLM}$ and evaluate the resulting model on SST-2, CoLA, QNLI, and RTE with 100 and 1000 training examples, respectively. The results are reported in Table \ref{tab:ablation}. As we can see, the results suffer severe deterioration by up to 7.5\% after removing $\mathcal{L}_{MLM}$, while removing $\mathcal{L}_{CL}$ only leads to a drop by up to 1.4\%. Although both $\mathcal{L}_{MLM}$ and $\mathcal{L}_{CL}$ contribute to the improvement of CMLM, MLM tends to play a more essential role. 
	
	\begin{table}[t]
		\centering
		\scalebox{0.75}{
    		\begin{tabular}{p{2.2cm}p{1.5cm}p{1.5cm}p{1.5cm}p{1.2cm}}
    			\hline
    			\multicolumn{1}{l}{} & \textbf{SST-2} & \textbf{CoLA} & \textbf{QNLI} & \textbf{RTE} \\ \hline
    			\textbf{Metric}      & acc            & mcc          & acc           & acc           \\ \hline
    			\multicolumn{5}{c}{data size = 100}                                                  \\ \hline
    			{CMLM}        & 0.8525         & \textbf{0.2663} & \textbf{0.6980} & 0.6147        \\
    			{w/ SimCLR}            & \textbf{0.8586}& 0.2511       & 0.6885        & \textbf{0.6355}        \\
    			w/o CL               & 0.8496         & 0.2626       & 0.6980        & 0.6095        \\
    			w/o MLM              & 0.8280         & 0.2492       & 0.6873        & 0.5913        \\
    			w/o CRM              & 0.8511         & 0.2621       & 0.6920        & 0.6242        \\ \hline
    			\multicolumn{5}{c}{data size = 1000}                                                 \\ \hline
    			{CMLM}        & 0.9023         & 0.4374 & \textbf{0.7719}    & 0.7610        \\
    			{w/ SimCLR}            & \textbf{0.9041} & \textbf{0.4446} & 0.7696     & \textbf{0.7732}        \\
    			w/o CL               & 0.9016         & 0.4362       & 0.7689        & 0.7524        \\
    			w/o MLM              & 0.8927         & 0.3983       & 0.7623        & 0.7039        \\
    			w/o CRM              & 0.9013         & 0.4434       & 0.7698        & 0.7654        \\ \hline
    		\end{tabular}
		}
		\caption{\label{tab:ablation}Results of ablation study for CMLM. \textit{w/ SimCLR} means replacing SimSiam with SimCLR, \textit{w/o CL} and \textit{w/o MLM} mean removing $\mathcal{L}_{CL}$ and $\mathcal{L}_{MLM}$ terms from $\mathcal{L}_{CMLM}$, respectively, and \textit{w/o CRM} means replacing CRM with dynamic random masking (DRM). }
	\end{table}
	
\subsection{Complementary Random Masking vs Dynamic Random Masking}
	We propose a complementary random masking (CRM) strategy to generate complementary masked sequences $T^k,\ k \in [1,K]$, based on $T^0$, which is generated by dynamic random masking (DRM). Here, we verify whether this complementary nature of $T^k$ benefits contrastive learning. We replace CRM in Equation \ref{eq:mask} by DRM, and conduct experiments on SST-2, CoLA, QNLI and RTE with 100 and 1000 training examples, respectively. As shown in Table \ref{tab:ablation}, CRM still surpasses DRM on all 8 tasks, with improvement by up to 1.6\%. The superiority of CRM mainly comes from fact that it avoids tokens to be masked in both $T^0$ and $T^k$.
	
\section{Conclusion}
	In this paper, we proposed a novel post-training objective, CMLM, for pre-trained language models in downstream few-shot scenes. CMLM attempts to combine both token-level and sequence-level contrastive learnings for more efficient domain transfer during post-training. For sentence-level contrastive learning, we developed a random masking strategy, CRM, to generate a pair of complementary masked sequences for an input sequence. Empirical results show that post-training with our CMLM outperforms other recent approaches on the GLUE tasks with 100 and 1000 labeled training examples, respectively. We also conducted extensive ablation studies and showed that both token-level and sequence-level contrastive learnings contribute to the results of CMLM, and that CRM achieves favorable sequence-level contrastive learning over the previous masking strategy. In future work, we will further investigate how token-level and sequence-level contrastive learnings affect domain transfer in post-training and explore more effective methods for sequence-level contrastive learning.
	
\section*{Acknowledgments}
The paper was supported by the Program for Guangdong Introducing Innovative and Entrepreneurial Teams (No.2017ZT07X355).
	
\bibliographystyle{acl_natbib}
\bibliography{acl2021}

\end{document}


\appendix

\noindent{\Large \textbf{Appendix}}
\section{Is Random Masking Good Enough for Data Augmentation?}
Recently, \newcite{fang2020cert} implement contrastive learning by using back-translation \cite{edunov-etal-2018-understanding} for data augmentation to generate sentence pairs. Their empirical results show that back-translation is better than easy data augmentation (EDA) \cite{wei-zou-2019-eda}, though it is more complicated. In this experiment, we implement back-translation with an online platform\footnote{http://api.fanyi.baidu.com/} for comparison. We wonder if our contrastive random masking (CRM) can be a better data augmentation method than EDA and back-translation, and whether different translation systems affect the results.
	
To this end, we replace back-translation in CSSL \cite{fang2020cert} with EDA, DRM, and CRM respectively. We also implement back-translation by MarianMT \cite{mariannmt} to evaluate the effect of different translation systems. We conduct experiments on SST-2, CoLA, QNLI, and RTE with data size of 100 and 1000, respectively. The results are reported in Table \ref{tab:mlm_good}.
	
As can be seen, CRM surpasses EDA, DRM, and back-translation with MarianMT when the data size is 100, while EDA and back-translation with MarianMT perform slightly better when the data size is 1000. CRM has shown superiority on sentence-level contrastive learning over DRM in previous sections. As for CRM and EDA, we assume their difference in performance comes from the degree of reformation between two augmented sequences: CRM generates two sequences with no structural change but only token replacement, where sentence-level similarity is much easier to be captured, while EDA might generate sequences with more obvious changes, which provides more information about data distribution but is more difficult to learn. As for back-translation, we find that the effectiveness of this method indeed depends on the selection of translation systems to certain extent, especially when the data size is 100. Therefore, we suspect random strategies like EDA and random masking are more suitable for few-shot scenes, which free a model from depending too much on the the selection of translation systems.
	
	\begin{table}[]
		\centering
				\scalebox{0.75}{
		\begin{tabular}{p{1.8cm}p{1.5cm}p{1.5cm}p{1.5cm}p{1.2cm}}
			\toprule
			\multicolumn{1}{l}{} & \textbf{SST-2} & \textbf{CoLA} & \textbf{QNLI} & \textbf{RTE} \\ \hline
			\textbf{Metric}      & acc            & mcc           & acc           & acc          \\ \hline
			\multicolumn{5}{c}{data size = 100}                                                  \\ \hline
			BACK                 & \textbf{0.8401}& 0.1719          & 0.6701         & 0.5532       \\
			BACK-M               & 0.8231         & 0.1520          & 0.6720         & 0.5463       \\
			EDA                  & 0.8299         & 0.1739          & 0.6685         & 0.5584       \\
			DRM                  & 0.8260         & 0.2466          & 0.6834         & 0.5905       \\
			{CRM}         & 0.8280         & \textbf{0.2492}& \textbf{0.6873} & \textbf{0.5913} \\ \hline
			\multicolumn{5}{c}{data size = 1000}                                                 \\ \hline
			BACK                 & 0.8993          & 0.4069          & 0.7660          & 0.7082       \\
			BACK-M               & 0.8995          & 0.4135          & \textbf{0.7765} & \textbf{0.7203}       \\
			EDA                  & \textbf{0.9014} & \textbf{0.4232} & 0.7746          & 0.6805      \\
			DRM                  & 0.8914          & 0.3890          & 0.7603          & 0.7025        \\
			{CRM}         & 0.8927          & 0.3983          & 0.7623          & 0.7039       \\ \bottomrule
		\end{tabular}
		}
		\caption{\label{tab:mlm_good} Results by replacing back-translation (BACK) in CSSL\cite{fang2020cert} with MarianMT (BACK-M), easy data augmentation (EDA), dynamic random masking (DRM), and contrastive random masking (CRM). Average results over 3 random seeds and 5 few-shot train sets are reported.}
	\end{table}

\bibliographystyle{acl_natbib}
\bibliography{acl2021}